\documentclass[10pt,twocolumn,letterpaper]{article}

\usepackage{iccv}              

\definecolor{iccvblue}{rgb}{0.21,0.49,0.74}
\usepackage[pagebackref,breaklinks,colorlinks,allcolors=iccvblue]{hyperref}
\usepackage{subcaption}
\usepackage{graphicx}
\usepackage{xcolor}
\usepackage[ruled,linesnumbered]{algorithm2e}

\def\paperID{13339} 
\def\confName{ICCV}
\def\confYear{2025}

\title{Decision PCR: Decision version of the Point Cloud Registration task}

\author{Yaojie Zhang, Tianlun Huang, Weijun Wang, Wei Feng\\
Shenzhen Institute of Advanced Technology, Chinese Academy of Sciences\\
Shenzhen, 518055, China\\
{\tt\small 1754133505@qq.com}
}

\begin{document}
\maketitle
\begin{abstract}

Low-overlap point cloud registration (PCR) remains a significant challenge in 3D vision. Traditional evaluation metrics, such as Maximum Inlier Count, become ineffective under extremely low inlier ratios. In this paper, we revisit the registration result evaluation problem and identify the Decision version of the PCR task as the fundamental problem. To address this Decision PCR task, we propose a data-driven approach. First, we construct a corresponding dataset based on the 3DMatch dataset. Then, a deep learning-based classifier is trained to reliably assess registration quality, overcoming the limitations of traditional metrics. To our knowledge, this is the first comprehensive study to address this task through a deep learning framework. We incorporate this classifier into standard PCR pipelines. When integrated with our approach, existing state-of-the-art PCR methods exhibit significantly enhanced registration performance. For example, combining our framework with GeoTransformer achieves a new SOTA registration recall of 86.97\% on the challenging 3DLoMatch benchmark.
Our method also demonstrates strong generalization capabilities on the unseen outdoor ETH dataset.

\end{abstract}    
\section{Introduction}
\label{sec:intro}

Point cloud registration (PCR) is a critical and foundational task in 3D computer vision. Currently, partial-to-partial registration remains challenging \cite{huang2021predator}, particularly when the overlap region between point clouds is small.  

Low-overlap scenarios severely challenge the model-fitting module in PCR methods. Most model-fitting algorithms follow a \textit{hypothesis generation and evaluation} paradigm. For evaluation, current state-of-the-art PCR methods\cite{yang2020teaser,chen2022sc2,yang2024mac} aim to identify 
transformations that maximize the consensus set of initial correspondences. However, under low-overlap conditions, the inlier rate drops to extremely low levels, frequently leading to registration failures.


To fundamentally address the low overlap challenge, two primary approaches emerge: (1) Developing more distinctive feature descriptors to improve the inlier rate, an active research topic with inherent challenges under low overlap condition; (2) Designing a more reliable evaluation criterion. Although recent efforts \cite{chen2023sc,xing2024efficient, yang2024mac,zhang2024svc} have attempted to modify the rule-based criteria using features, overlap priors or additional constraints, no prior work has treated this as an independent task or conducted a systematic investigation. To bridge this gap, this paper formally proposes the Decision version of the PCR problem (Decision PCR).

\begin{figure}[!t]
  \centering
  \fcolorbox{white}{white}{\includegraphics[width=\linewidth]{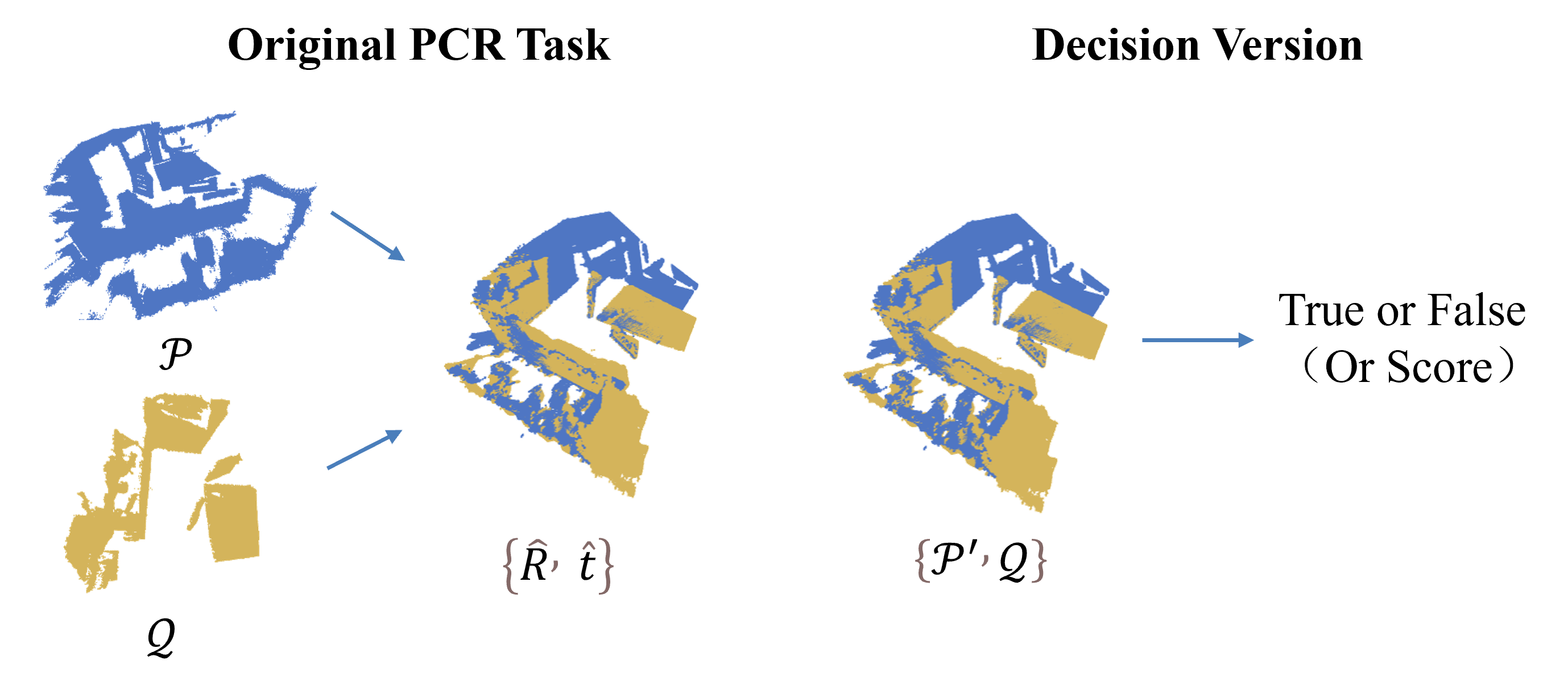}}
  \caption{The decision version of the PCR task aims to evaluate the quality of the merged point cloud.}
  \label{fig:decision_version}
\end{figure}

The objective of Decision PCR is: Given two partially overlapping point clouds and multiple pose transformation hypotheses, the task aims to develop a model capable of accurately classifying these hypotheses as correct or incorrect transformations. Fig.\ref{fig:decision_version} provides an intuitive illustration of this task. Its key distinction lies in requiring only a correctness judgment for a given transformation, rather than directly estimating the transformation. However, no existing rule-based metric can reliably address this problem. In this work, we address it for the first time using a data-driven deep learning framework, which necessitates two components: (1) Constructing a specialized dataset for this task, and (2) Developing a corresponding training framework.

\textbf{Why Study the Decision PCR Task as an Independent Research Problem?} The Decision PCR task exhibits multiple advantages over the original PCR task:
\begin{itemize}
    \item Theoretical Basis: For any reliably solvable problem, result validation is essential. The decision-version task therefore forms the theoretical foundation of the original PCR task and must be addressed first.
    \item Evaluation Layer: It adds a reliable layer to existing PCR frameworks for validating alignment quality. Current PCR frameworks typically assume the registration result is optimal, but lack an assessment of the transformation correctness probability.
    \item Data Scalability: The inherent properties of this task enable scaling of existing PCR datasets. By incorporating knowledge from both correct and incorrect alignments, it effectively expands usable training data.
    \item Practical Simplicity: This task is structurally simpler and more intuitive than the original PCR. It can be naturally formulated as a point cloud binary classification problem, enabling direct deployment of advanced models from related domains.
\end{itemize}

To demonstrate its importance, we integrate the Decision PCR model with existing PCR methods as an evaluation layer. This decision PCR based framework addresses a critical limitation in traditional PCR methods: the lack of assessment for the registration result. We will demonstrate the advantages of incorporating this model in the experiments section. In summary, our contributions are:
\begin{enumerate}
    \item We conduct a systematic investigation into the decision version of the PCR problem and, for the first time, propose a data-driven approach to address it.
    \item Building upon our proposed Decision PCR model, we develop a reliable PCR method that not only addresses the limitations of initial correspondence quality in evaluation but also provides uncertainty-aware score for the registration result.
    \item Extensive experiments on multiple PCR benchmarks demonstrate the state-of-the-art performance of our method. Notably, our approach achieves superior results on the unseen outdoor ETH dataset, highlighting the robust generalization capability of the Decision PCR model. 
\end{enumerate}

\section{Related Work}
\label{sec:related_work}

\subsection{3D Feature Matching}
3D feature matching aims to generate precise initial correspondences based on representative feature descriptors. Deep learning techniques have significantly advanced the learning of 3D local descriptors, surpassing traditional hand-crafted approaches. FCGF \cite{choy2019fully} computes features in a single pass through a fully convolutional neural network. Some studies focus on rotation-equivariant descriptors like D3feat \cite{bai2020d3feat}, SpinNet \cite{ao2021spinnet} and RoReg \cite{wang2023roreg}. For low-overlap PCR challenges, some studies propose specialized modules or matching strategies to improve the inlier rates.PREDATOR \cite{huang2021predator} employs an attention mechanism to pre-detect overlapping regions. CoFiNet \cite{yu2021cofinet} adopts a coarse-to-fine matching strategy building upon PREDATOR. Extending this pipeline, GeoTransformer \cite{qin2022geometric} introduces a novel position encoding technique and learns geometric features for superpoints. PEAL\cite{yu2023peal} further incorporates the overlap prior of the GeoTransformer by utilizing a one-way attention mechanism, thus improving the inlier rate. The core idea of these methods is to first predict overlapping regions and then perform local point-to-point matching within these regions to improve the inlier rate of initial correspondences. However, this paradigm critically depends on the accuracy of overlapping region prediction---a task inherently challenging under low-overlap conditions.





\subsection{Model Fitting}
Given initial correspondences, model fitting aims to remove outliers and estimate the optimal pose transformation. The most widely used approach, RANSAC \cite{fischler1981random}, employs a generate-and-verify pipeline for robust outlier removal. However, RANSAC and its variants \cite{barath2018graph,quan2020compatibility,barath2019magsac,chum2005matching} often suffer from inefficiency under low inlier ratios. Global optimal approaches can also achieve outlier robustness registration. For example, Go-ICP \cite{yang2015go} utilizes a branch-and-bound (BnB) scheme for globally optimal registration, while FGR \cite{zhou2016fast} applies the Geman-McClure cost function and estimates the model through graduated non-convexity optimization. Besides, the spatial compatibility method is also widely applied in point cloud registration. Teaser \cite{yang2020teaser} introduces a graph-theoretic framework for robust data association. PointDSC \cite{bai2021pointdsc} develops a spatial consistency-based non-local module and a neural spectral matching mechanism to accelerate model generation and selection. DHVR \cite{lee2021deep} generates hypotheses for deep Hough voting using SC-validated tuples. SC2 \cite{chen2022sc2} proposes a second-order spatial compatibility measure, enabling more distinctive clustering compared to the original metric. MAC \cite{zhang20233d} relaxes the maximum clique constraint to a maximal clique constraint, allowing for richer local graph information extraction.



\subsection{Metrics for Model fitting.}
\begin{figure*}[!ht]
  \centering
   \includegraphics[width=0.95\textwidth]{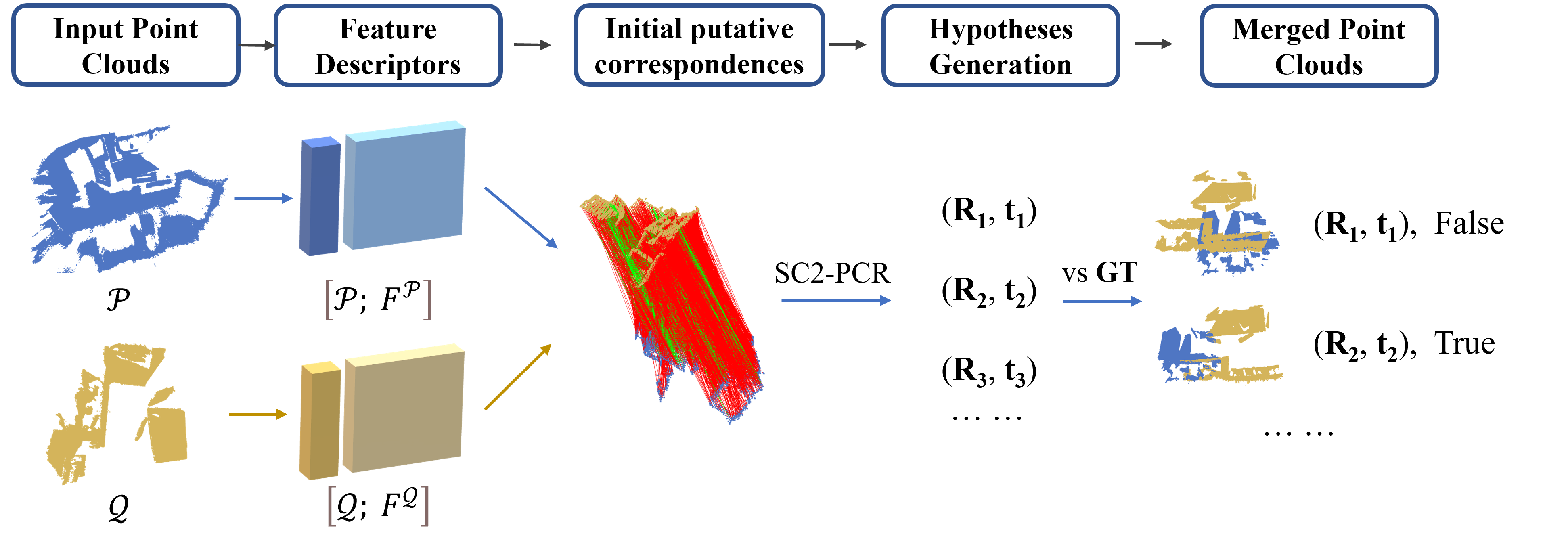}

   \caption{The pipeline of dataset construction. Existing PCR method \cite{chen2022sc2} is used to generate challenging wrong transformation candidates. The output is merged point clouds with corresponding labels.}
   \label{fig:dataset_generating}
\end{figure*}

Model fitting typically selects the "optimal" model based on the best score according to a specific evaluation metric, such as the widely used Maximum Inlier Count (MIC), which serves as the optimization objective. Recent works have improved evaluation metrics by incorporating prior information or additional constraints, yielding promising results. For instance, SC2++ \cite{chen2023sc} combines feature descriptor information with spatial consistency to propose the Feature and Spatial Consistency Constrained Truncated Chamfer Distance (FS-TCD) metric. MAC-OP \cite{yang2024mac} leverages overlap priors to enhance metric robustness. VDIR \cite{xing2024efficient} identifies the optimal transformation by minimizing viewpoint deviation distance. Alternatively, SVC \cite{zhang2024svc} applies the Sight View Constraint to eliminate definitively incorrect transformations, thereby narrowing the candidate range.

These methods modify rule-based metrics with additional features or overlap prior information, which remain heavily dependent on initial correspondence quality. In contrast, our work proposes a novel deep learning-based approach that directly processes raw point clouds, effectively mitigating the impact of initial correspondences.

\section{Method}

\subsection{Overview}
To address the Decision PCR problem and integrate it into the original PCR framework, three key steps are required: (1) Constructing a dedicated dataset for the Decision PCR task (Sec.\ref{sec:dataset_construction}), (2)Training a classifier tailored for Decision PCR (Sec.\ref{sec:model_training}), and (3) Integrating the classifier into existing PCR pipelines (Sec.\ref{sec:integrate_PCR}).
\subsection{Dataset Construction}\label{sec:dataset_construction}
While established datasets \cite{geiger2013vision,zeng20173dmatch,wu20153d} exist for the original PCR task, they require adaptation to meet Decision PCR task requirements. Our objective is to generate challenging incorrect transformations for the Decision PCR task. To be more specific, given a target point cloud $\mathcal{Q} = \{ \mathbf{q_j} \in \mathbb{R}^3\ |\ j = 1, ..., n \}$ and a source point cloud $\mathcal{P} = \{ \mathbf{p_i} \in \mathbb{R}^3\ |\ i = 1, ..., m \}$ with Ground Truth transformation $(\mathbf{{R}^*},  \mathbf{t}^*)$, this procedure aims to get challenging wrong transformation $(\mathbf{\hat{R}},  \mathbf{\hat{t}})$ first, then create the merged point clouds $\{\mathcal{P'},\mathcal{Q}\}$ where $\mathcal{P'}$ denotes the transformed source point cloud:
\begin{equation}
    \mathcal{P'} = \{ \mathbf{\hat{R}}\mathbf{p_i} +\mathbf{\hat{t}}|\  \mathbf{p_i} \in \mathcal{P}\}.
    \label{eq:transform_points}
\end{equation}

Please note that random incorrect transformations are insufficient as they are easily distinguishable from correct ones.  We therefore design a framework to generate challenging wrong transformations as shown in Fig.\ref{fig:dataset_generating}. 
Using all pairs with over 10\% overlap rate in train set of 3DMatch, we construct the Decision PCR dataset as follows: First, downsample input point clouds and extract feature descriptors to generate initial correspondences. Since the goal of this step is to generate wrong transformations, we intentionally select the FPFH descriptor, which has limited performance on this dataset. Next, the SC2 \cite{chen2022sc2} PCR method is leveraged to generate hundreds of transformation hypotheses. These hypotheses are ranked in descending order based on the inlier count metric, which serves as a proxy for estimating the size of overlap regions. Finally, wrong transformations are selected by jointly evaluating overlap ratios and transformation errors relative to ground truth.

Typical wrong transformations are categorized into four types: large-overlap cases, small-overlap cases, large-transformation-error cases, and small-transformation-error cases. The final dataset contains \textbf{26,605} correct and \textbf{97,794} incorrect merged point clouds. Each merged point cloud $\{\mathcal{P'},\mathcal{Q}\}$ includes explicit source/target tags for every point. Which means $\mathbf{p}=(x,y,z,tag)$, for $\mathbf{p} \in \{\mathcal{P'},\mathcal{Q}\}$.

\subsection{Model training}\label{sec:model_training}
\begin{figure}[!b]
  \centering
   \includegraphics[width=0.9\linewidth]{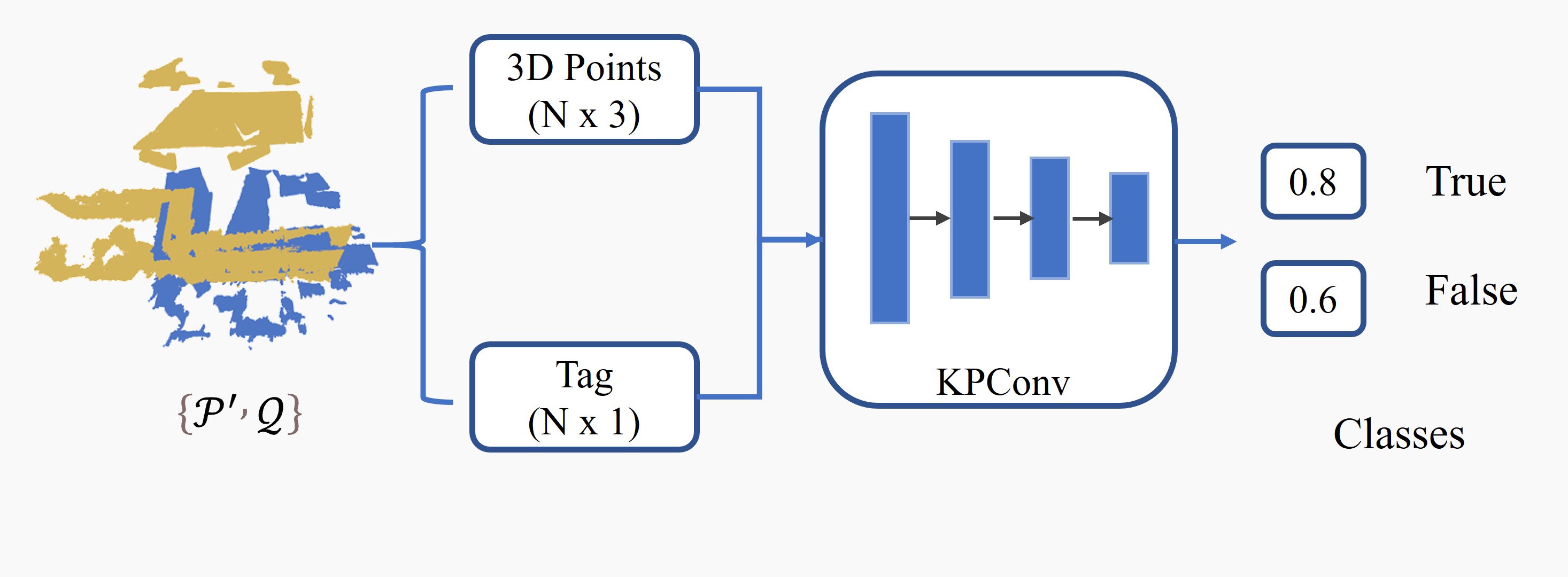}

   \caption{The pipeline of model training.}
   \label{fig:model_training}
\end{figure}

Given the merged point cloud $\{\mathcal{P'},\mathcal{Q}\}$, this section aims to train a model $F(*)$ to score it. 
\begin{equation}
    Score = F(\{\mathcal{P'},\mathcal{Q}\})
\end{equation}

\begin{figure*}[h]
  \centering
   \includegraphics[width=0.95\textwidth]{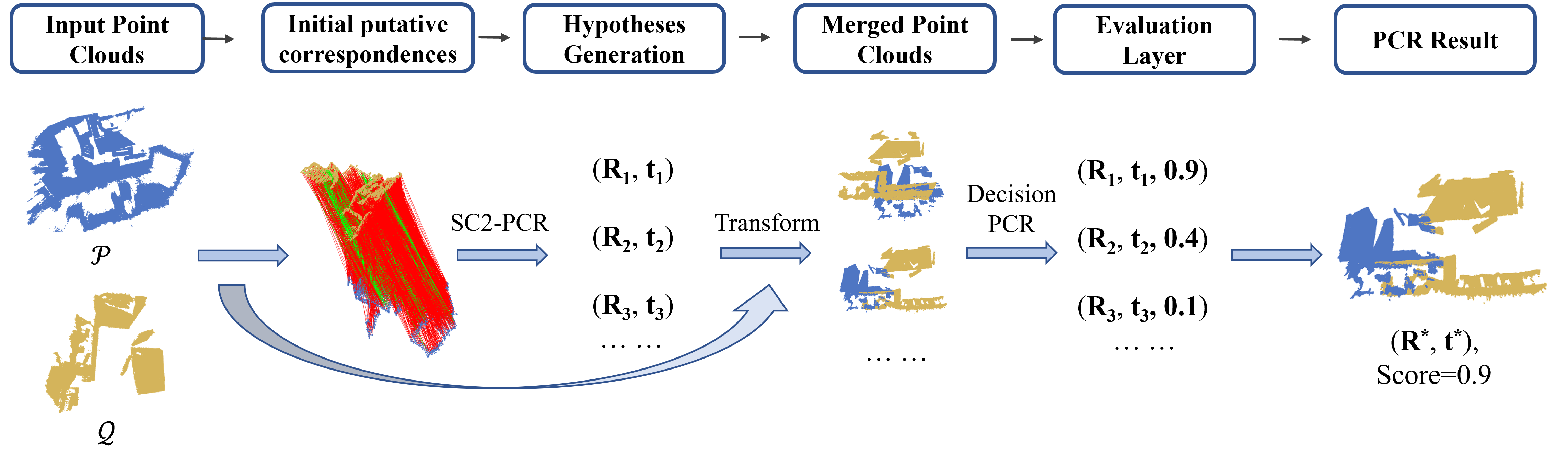}
   \caption{PCR pipeline based on Decision PCR model. (1) Hypotheses and input point clouds are used to generate merged point clouds. (2) The Decision PCR model is employed to evaluate scores. (3) The optimal transformation is selected using Algorithm \ref{alg:evaluation}.}
   \label{fig:integrate_PCR}
\end{figure*}

In this work, we frame the Decision PCR task as a binary point cloud classification problem (Correct vs. Wrong transformations), enabling the use of established classification networks like PointNet \cite{aoki2019pointnetlk} and KPConv \cite{thomas2019kpconv}. 

The obtained dataset is split into an 80\% training set and 20\% validation set. We adopt the KPCNN architecture of KPConv \cite{thomas2019kpconv} for classification, with the pipeline illustrated in Fig. \ref{fig:model_training}. The model takes merged point clouds $\{\mathcal{P'},\mathcal{Q}\}$ along with associated tags for each point as input. The output is a two-dimensional vector $\mathbf{v}=(v_t,v_f)$, where the predicted label corresponds to the component with the larger numerical value. For all other training parameters and the model framework, we adhere to the configurations provided on the official KPConv\footnote{https://github.com/HuguesTHOMAS/KPConv-PyTorch}. Consequently, we obtain a classifier capable of evaluating merged point clouds. 
For the obtained two-dimensional vector$\mathbf{v}$, it can also be processed through a Softmax function to derive a corresponding $Score$ that represents the probability of the transformation being correct.
\begin{equation}
    Score = Softmax(v_t)=\frac{e^{v_t}}{e^{v_t}+e^{v_f}}
\end{equation}

On the validation set, the model achieves approximately 96.8\% average accuracy, with the detailed results presented in the Tab.\ref{tab:decision_model}. Notably, incorporating point tags significantly improves performance, distinguishing this task from general point cloud classification. With this enhanced model, we are now able to evaluate the input merged point cloud and estimate the probability of the transformation being correct. To further highlight its importance in the context of the original PCR task, we need to integrate this model into the existing PCR pipeline.

\begin{table}[!htbp]
    \raggedright
    \small
    \begin{tabular}{cccccc}
        \toprule
         & Correct Acc. & Wrong Acc. & Avg. Acc. \\
         & (\%) & (\%) & (\%) \\
        \midrule
        without tag &88.2 & 92.9 & 91.9\\
        with tag & 91.7 & 98.2 & 96.8 \\ 
        \bottomrule
    \end{tabular}
    \caption{Performance on validation set.}
    \label{tab:decision_model}
\end{table}


\subsection{Integrating the Decision Model into Existing PCR pipeline}\label{sec:integrate_PCR}

Most advanced PCR methods adopt a \textit{hypotheses generation and evaluation} paradigm. The proposed Decision PCR model can seamlessly integrate into the existing PCR pipeline as an evaluation layer. The illustration of the new pipeline combined with Decision PCR model is shown in the Fig.\ref{fig:integrate_PCR}. Traditional PCR method can only select the "optimal" transformation under a specific evaluation metric, while our model can assign a confidence score to the obtained results.

The detailed description of the Fig.\ref{fig:integrate_PCR} is as follows. First, the existing PCR method SC2 \cite{chen2022sc2} is employed to generate transformation hypotheses. Then we apply these transformations to the input point clouds using Eq.\ref{eq:transform_points} to generate merged point clouds. By leveraging the original point cloud information in this step, our evaluation layer is rendered independent of the quality of the initial correspondences. Subsequently, merged point clouds are fed into the Decision PCR model to obtain corresponding scores, which are then used to select the optimal transformation. By utilizing the evaluation layer based on our Decision PCR model, we can robustly select the optimal transformation with a confidence score.

However, the aforementioned description implies a critical issue: SC2\cite{chen2022sc2} may generate hundreds of hypotheses, significantly compromising computational efficiency. To address this issue, we employ two strategies: 1) utilizing the sight view constraint \cite{zhang2024svc} to eliminate some obvious wrong transformations thereby reducing the number of hypotheses, and 2) performing early truncation based on confidence scores. The detailed algorithm is shown in the Alg.\ref{alg:evaluation}.

\begin{algorithm}[!b]
    \caption{Evaluation algorithm based on Decision PCR model}
  \label{alg:evaluation}
  \KwIn{Input Point clouds $\mathcal{P}$,$\mathcal{Q}$; \\
  Transformation Hypotheses $\mathcal{T}:\{T_0, T_1,..., T_K\}$}
   \tcp{Generated by SC2 \cite{chen2022sc2}.}
  \KwOut{Optimal transformations $T^*$, $Score^*$}
  \tcp{Generate Filtered transformations $\mathcal{T}_f:\{\}$ using SVC \cite{zhang2024svc}.}
  \textit{While} (count $<$ m):\\
    \quad if $SVC(\mathcal{P},\mathcal{Q},T_i)$ is True:\\
    \quad \quad $\mathcal{T}_f \leftarrow T_i$\\
    \quad \quad Count = Count + 1\\
    \quad i = i + 1\\
  \tcp{Score transformations using the Decision PCR model $F(*)$.}  
  \textit{for\ every} $T_j \in \mathcal{T}_f$:\\
  \quad Generate $\{\mathcal{P'},\mathcal{Q}\}$ according to Eq.\ref{eq:transform_points} \\
  \quad $Score = F(\{\mathcal{P'},\mathcal{Q}\})$\\
  \tcp{Truncation according to $threshold$}
  \quad \textit{if} $Score > threshold$:\\
  \quad \quad  $T^*=T_j$,\ $Score^*=Score$\\
  \quad \quad  Break the loop\\
  \quad \textit{elif} $Score > Score^*$:\\
  \quad \quad $T^*=T_j$,\ $Score^*=Score$\\
return $T^*$,  $Score^*$   
  
\end{algorithm}





\section{Experiments}
\label{sec:experiments}

\subsection{Datasets and Experimental Setup}
\subsubsection{Datasets and Evaluation Criteria}
\textbf{3DMatch \& 3DLoMatch:} 3DMatch \cite{zeng20173dmatch} is a widely used indoor RGB-D dataset, consisting of 62 scenes, divided into 46 scenes for training, 8 for validation, and 8 for testing. Point clouds are reconstructed from RGB-D frames, preserving real-world sensor noise. We train the Decision PCR model on the training set. For PCR evaluation, we follow PREDATOR \cite{huang2021predator} to partition the test set into 3DMatch (1,623 pairs) and 3DLoMatch (1,781 pairs).

\textbf{ETH}: ETH\cite{pomerleau2012challenging} is an outdoor dataset used exclusively for testing, consisting of 713 pairs derived from 132 point clouds across 4 scenes, which were collected using a laser scanner. In this work, it is mainly used to evaluate the generalization performance of our approach.

\textbf{Evaluation Criteria}: Following previous works \cite{zhang20233d}, the primary indicator is
registration recall (RR) under an error threshold. For {3DMatch, 3DLoMatch, and ETH datasets}, the threshold is set to (15 deg, 30 cm). Transformation errors are quantified using relative rotation error (RE) and L2 translation error (TE):

\begin{equation}
RE(\mathbf{\hat{R}})=acos (\frac{trace(\mathbf{\hat{R}^T}\mathbf{R^*})-1}{2}),\ TE(\mathbf{\hat{t}})=||\mathbf{\hat{t}}-\mathbf{t^*}||_2.
  \label{eq:error_metric}
\end{equation}
Here $\mathbf{R^*}$ and $\mathbf{t^*}$ denote the ground-truth rotation and translation. 

\subsubsection{Implementation Details}
For the Decision PCR task, the model is trained on a single NVIDIA GeForce RTX 3090 GPU with a batch size of 32 for 2,000 iterations. The voxel downsampling size of the point clouds used for model training is set to 5 cm. We use the point tag as an input feature for the normal channel. For other parameters and the model structure, we follow the original KPConv \cite{thomas2019kpconv}. For the parameters in Alg.\ref{alg:evaluation}, the score threshold is set to 0.6, and $m$ is set to 100. For the SC2 \cite{chen2022sc2} and SVC \cite{zhang2024svc} module, we remain the recommended parameters.

\begin{table*}[!htbp]
  \centering
  \small
  \begin{tabular}{lccccccccccccc}
    \toprule
    & & \multicolumn{2}{c}{3DMatch FPFH} & \multicolumn{2}{c}{3DMatch FCGF} & \multicolumn{2}{c}{3DLoMatch FPFH} & \multicolumn{2}{c}{3DLoMatch FCGF} & \\
    \cmidrule(lr){3-4} \cmidrule(lr){5-6} \cmidrule(lr){7-8} \cmidrule(lr){9-10}
     & &  & RE(deg) &  & RE(deg) &  & RE(deg) &  & RE(deg) & Time\\
     & & RR(\%) & /TE(cm) & RR(\%) & /TE(cm) & RR(\%) & /TE(cm) & RR(\%) & /TE(cm) & (s)\\
    \midrule
    \multicolumn{10}{l}{\textbf{Deep Learned}} \\
    DGR \cite{choy2020deep}& & 32.84 & 2.45\ /\ 7.53 & 88.85 & 2.28\ /\ 7.02 & 19.88 & 5.07\ /\ 13.53 & 43.80 & 4.17\ /\ 10.82 & 1.53\\
    DHVR \cite{lee2021deep}& & 67.10 & 2.78\ /\ 7.84 & 91.93 & 2.25\ /\ 7.08 & - & -\ /\ - & 54.41 & 4.14\ /\ 12.56 & 3.92\\
    PointDSC \cite{bai2021pointdsc} & & 77.39 & {2.05\ /\ 6.43} & 92.85 & 2.05\ /\ {6.50} & 27.74 & 4.11\ /\ 10.45 & 55.36 & 3.79\ /\ 10.37 & \textbf{0.10} \\
    \midrule
    \multicolumn{10}{l}{\textbf{Traditional}} \\
    RANSAC-1M \cite{fischler1981random} & & 65.29 & 3.52\ /\ 10.98 & 89.62 & 2.50\ /\ 7.55 & 15.34 & 6.05\ /\ 13.74 & 46.38 & 5.00\ /\ 13.11 & 0.97\\
    GC-RANSAC \cite{barath2018graph}& & 71.97 & 2.43\ /\ 7.03 & 89.53 & 2.25\ /\ 6.93 & 17.46 & 4.43\ /\ 10.75 & 41.83 & 3.90\ /\ 10.44 & 0.55\\
    TEASER \cite{yang2020teaser}& & 75.79 & 2.43\ /\ 7.24 & 87.62 & 2.38\ /\ 7.44 & 25.88 & 4.83\ /\ 11.71 & 42.22 & 4.65\ /\ 12.07 & 0.07\\
    FGR \cite{zhou2016fast}& & 40.91 & 4.96\ /\ 10.25 & 78.93 & 2.90\ /\ 8.41 & - & -\ /\ - & 19.99 & 5.28\ /\ 12.98 & 0.89\\
    SC2 \cite{chen2022sc2} (Baseline) & & 83.26 & {2.09}\ /\ 6.66 & 93.16 & 2.09\ /\ 6.51 & 38.46 & 4.04\ /\ 10.32 & 58.62 & 3.79\ /\ 10.37 & 0.11\\ 	 			 
    SC2++ \cite{chen2023sc}& & \underline{87.18} & 2.10\ /\ {6.64} & \underline{94.15} & 2.04\ /\ {6.50} & 41.27 & {3.86\ /\ 10.06}  & 61.15 & {3.72}\ /\ 10.56 & 0.28\\
    {VBReg} \cite{jiang2023robust} & & 82.75 & 2.14\ /\ 6.77 & 93.53 & {2.04}\ /\ {6.49} & - & -\ /\ - & 58.30 & -\ /\ - & 0.22\\
    MAC \cite{zhang20233d}& & 83.92 & 2.11\ /\ 6.79 & 93.72 & {2.03}\ /\ 6.53 & 41.27 & 4.06\ /\ 10.64 & 60.08 & {3.75}\ /\ 10.60 & 3.26\\
    MAC-OP \cite{yang2024mac} & & 84.78 & 2.29\ /\ 6.40 & 92.79 & 2.17\ /\ \textbf{6.11} & \textbf{46.77} & 3.79\ /\ 9.93 & \underline{62.32} & {3.66}\ /\ 9.78 & 1.45\\
    MAC-OP$\diamond$ \cite{yang2024mac} & & 87.55 & 2.34\ /\ 6.48 & 94.95 & 2.14\ /\ 6.03 & 54.69 & 4.01\ /\ 10.18 & 73.05 & {3.77}\ /\ 9.94 & 1.45\\
    \midrule
    Ours & & \textbf{88.29} & \textbf{1.72}\ /\ \textbf{6.38} & \textbf{95.13} & \textbf{1.68}\ /\ 6.37 & \underline{45.54} & \textbf{3.04}\ /\ \textbf{9.31} & \textbf{71.59} & \textbf{3.01}\ /\ \textbf{9.18} & 2.74\\
    \bottomrule
  \end{tabular}
  \caption{Quantitative Results on 3DMatch \& 3DLoMatch dataset. MAC-OP$\diamond$ utilizes ground-truth overlap score, for fair comparison, it is excluded from ranked comparisons.}
  \label{tab:results_on_3DMatch}
\end{table*}

\subsection{Results on Indoor Scenes}
\subsubsection{Baseline Methods}
We conduct extensive comparisons on 3DMatch \& 3DLoMatch benchmarks. Both deep-learned and traditional PCR methods are considered, including DGR \cite{choy2020deep}, DHVR \cite{lee2021deep}, PointDSC \cite{bai2021pointdsc}, SM \cite{leordeanu2005spectral}, RANSAC \cite{fischler1981random}, GC-RANSAC \cite{barath2018graph}, 
TEASER \cite{yang2020teaser}, FGR \cite{zhou2016fast}, SC2 \cite{chen2022sc2},  VBReg \cite{jiang2023robust}, SC2++ \cite{chen2023sc}, MAC \cite{zhang20233d} and MAC-OP \cite{yang2024mac}. To ensure a fair comparison, we use the same descriptors (including FPFH \cite{rusu2009fast} and FCGF\cite{choy2019fully}) and identical initial correspondence counts. Note that MAC-OP$\diamond$ utilizes Ground-Truth overlap information, representing the theoretical upper bound of MAC, and thus it is excluded from ranked comparisons. Quantitative results
are shown in Tab.\ref{tab:results_on_3DMatch}.

\subsubsection{Results on 3DMatch \& 3DLoMatch}
As shown in Tab.\ref{tab:results_on_3DMatch}, our method achieves best registration results on both 3DMatch and 3DLoMatch datasets compared to all other PCR methods. Compared to the baseline method SC2, our approach further boost its registration recall from 83.26\% \& 93.16\% to \textbf{88.29\%} \& \textbf{95.13\%} on 3DMatch. On the more challenging 3DLoMatch benchmark, the improvement is even more pronounced, increasing from 38.46\% \& 58.62\% to \textbf{45.54\%} \& \textbf{71.59\%}, respectively. Even when directly compared to MAC-OP$\diamond$, our method outperforms it on 3DMatch and achieves comparable RR on 3DLoMatch using FCGF descriptor. For the MAC-OP$\diamond$ method, despite leveraging Ground-Truth (GT) overlap information, its core remains a correspondence-based metric, which inherently depends on the quality of the initial correspondences. In contrast, our global point cloud evaluation paradigm eliminates this dependency.


\subsubsection{Combined with deep learned descriptors}

\begin{table}[!b]
  \centering
  \begin{tabular}{lccc}
    \toprule
     & \multicolumn{3}{c}{3DLoMatch} \\
     \cmidrule(lr){2-4}
    Method & RR(\%) & RE(deg)& TE(cm) \\
    \midrule
    \multicolumn{4}{l}{\textit{PREDATOR}\cite{huang2021predator}}\\
    SC2 & 69.68 & 3.46 & 9.66\\
    MAC & 70.90 & - & - \\
    MAC-OP & 69.70 & - & -\\
    MAC-OP$\diamond$ & 77.50 & - & -\\
    Ours & \textbf{78.27} & 2.98 & 9.24\\ 
    \midrule
    \multicolumn{4}{l}{\textit{GeoTransformer}\cite{qin2022geometric}}\\
    SC2 & 78.11 & 3.01 & 8.69 \\
    MAC & 78.90 & - & - \\
    MAC-OP & 77.80 & - & - \\
    MAC-OP$\diamond$ & 84.00 & - & - \\
    Ours & \textbf{86.97} & 2.95 &9.23 \\
    \midrule
    \multicolumn{4}{l}{\textit{PEAL}\cite{yu2023peal}}\\
    SC2 & 83.10 & 2.85 & 8.35 \\
    MAC & 83.30 & - & - \\
    MAC-OP & 82.40 & - & - \\
    MAC-OP$\diamond$ & 89.10 & - & - \\
    Ours & \textbf{86.13} & 2.86 &9.14 \\
    \bottomrule
  \end{tabular}
  \caption{{Comparison of different PCR Methods on ETH dataset.}}
  \label{tab:geotransformer_comparison}
\end{table}

Our approach can seamlessly integrate with other advanced deep-learned descriptors. In this section, we combine with representative descriptors such as PREDATOR\cite{huang2021predator}, GeoTransformer\cite{qin2022geometric}, and PEAL\cite{yu2023peal} on the challenging 3DLoMatch dataset. As the most advanced PCR method to date, MAC-OP\cite{yang2024mac} is listed as baseline for comparison.

As shown in Table \ref{tab:results_on_3DMatch}, our method surpasses both MAC and MAC-OP in RR across all descriptors. For PREDATOR and GeoTransformer, our results rival MAC-OP$\diamond$ (which uses ground-truth information), demonstrating superior inherent evaluation capability.

\textbf{Is using overlap prior a good approach?} Both most advanced modules PEAL\cite{yu2023peal} and MAC-OP \cite{yang2024mac} utilize overlap prior information for performance boosting. However, in our view, leveraging overlap prior information more as an expedient approach than a fundamental enhancement. For example, MAC-OP integrates PREDATOR’s overlap prior but exhibits paradoxical behavior: it improves weak descriptors (FPFH/FCGF) yet degrades performance on stronger ones (GeoTransformer/PEAL). For PEAL, although integrating the overlap prior enhances its inlier rate relative to GeoTransformer—thereby improving compatibility with other PCR methods—this modification fails to deliver fundamental improvements when applied to our method, which exhibits stronger inherent evaluation capability. These observations will be further substantiated in the comparative experiments presented in the following section. 


\subsubsection{Score the Registration Result}
\begin{figure}[!h]
  \centering
   \includegraphics[width=0.8\linewidth]{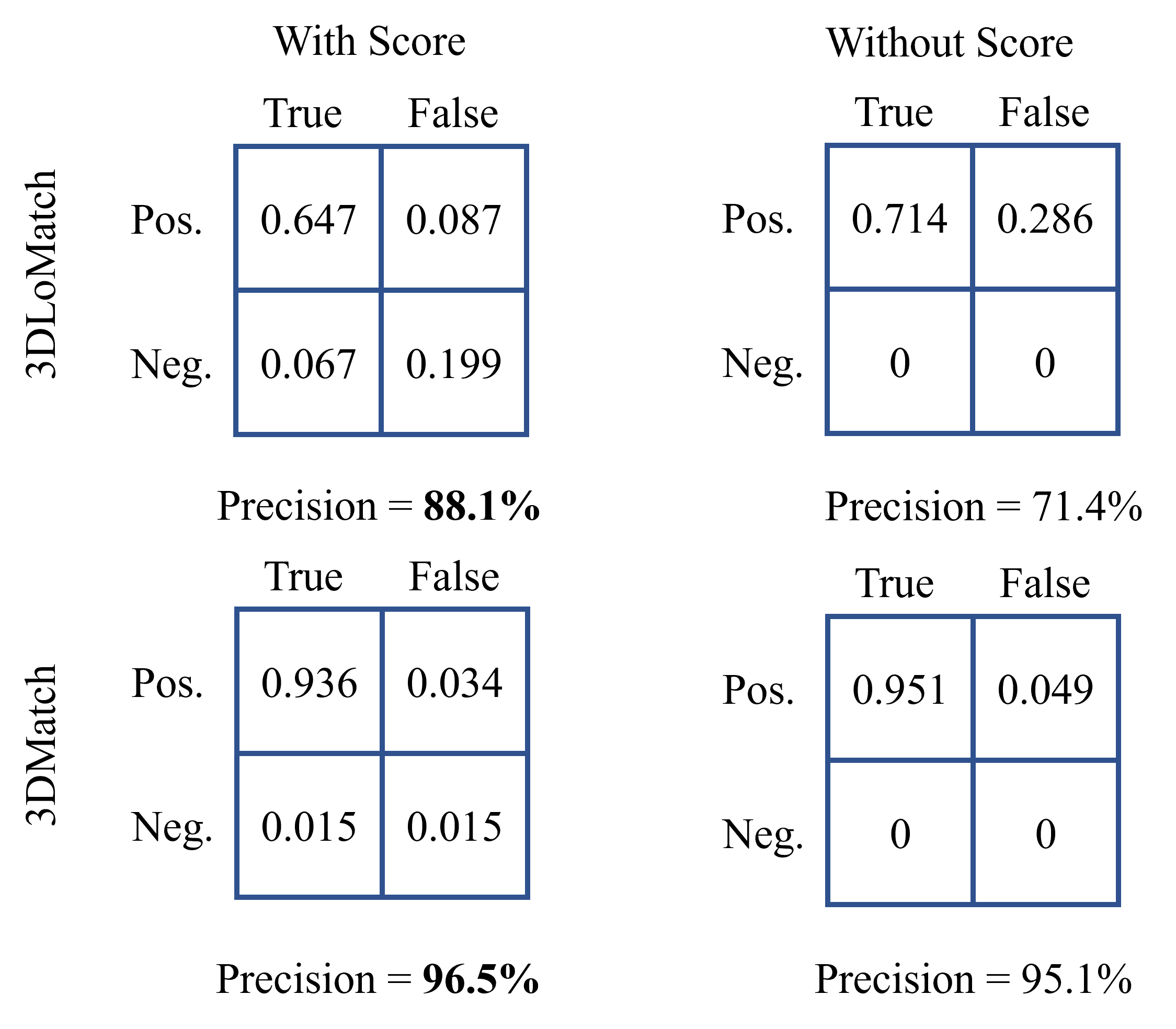}

   \caption{Confusion matrix of FCGF registration result on 3DLoMatch.}
   \label{fig:confusion_matrix}
\end{figure}

Traditional PCR methods focus solely on Registration Recall, implicitly assuming that the obtained results are optimal. While our approach can evaluate the quality of the registration recall by assigning a score via the Decision PCR model. As shown in the confusion matrix (Fig. \ref{fig:confusion_matrix}), we reclassify results using a score threshold of 0.5. For example, in the case of FCGF on the 3DLoMatch dataset, the probability of correct positive transformations increases from 71.4\% (baseline) to \textbf{88.1\%} using our model. This indicates that decision PCR model can significantly enhance the reliability of the existing PCR pipeline.

\subsection{Generalization Experiments}
To validate the generalization capability of the Decision PCR model, we directly applied our method to the outdoor ETH\cite{pomerleau2012challenging} dataset. Following the MAC\cite{yang2024mac} experimental protocol, we employed FPFH\cite{rusu2009fast}, FCGF\cite{choy2019fully}, SpinNet\cite{ao2021spinnet}, and GeoTransformer\cite{qin2022geometric} to generate initial correspondences. Deep-learning-based descriptors are also pre-trained on 3DMatch and evaluated on ETH without fine-tuning.

\subsubsection{Scale the merged point clouds.}
\begin{figure}[!hb]
  \centering
   \includegraphics[width=0.9\linewidth]{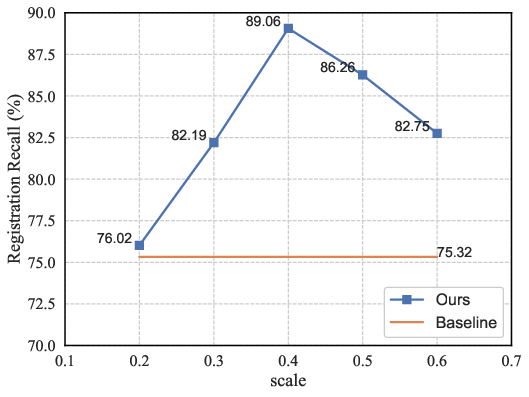}
   \caption{RR vs scale on ETH dataset with FCGF.}
   \label{fig:RR_scale}
\end{figure}

The KPConv-trained Decision PCR model is sensitive to the voxel downsampling size. For outdoor scenes, the voxel size is generally set to 30 cm. Maintaining indoor-scale voxels (5 cm) on ETH drastically increases point cloud density and computational overhead. We resolve this by simply scaling the merged point clouds while preserving outdoor sampling sizes. Under this condition, the $Score$ can be computed as:
\begin{equation}
    Score = F(scale*\{\mathcal{P'},\mathcal{Q}\})
\end{equation}

We set the $m=30$ and $threshold=0.6$ for the Alg.\ref{alg:evaluation}. By varying the $scale$ parameter, we establish the relationship between Registration Recall (RR) and $scale$, as depicted in the Fig.\ref{fig:RR_scale}. Baseline performance (without Decision PCR) serves as reference. The optimal RR occurs at $scale = 0.4$, which is adopted for the following experiment.

\subsubsection{Results on the ETH dataset}
To be consistency with the indoor setting, we set the $m=100$ and $threshold=0.6$. The detailed registration results is shown in the Tab.\ref{tab:ETH_performance}. It is evident that our method significantly outperforms all other approaches across all descriptors, even for the MAC using Ground Truth overlap prior. This result is consistent with the performance on the 3DMatch dataset, likely due to our model’s enhanced evaluation capability in moderate-overlap scenarios (both ETH and 3DMatch pairs have over 30\% overlap) compared to MAC-OP$\diamond$.

\begin{table}[!htbp]
\centering
\footnotesize
\begin{tabular}{lccccc}
\toprule
 & \multicolumn{2}{c}{\textbf{Gazebo}} & \multicolumn{2}{c}{\textbf{Wood}} & \textbf{Avg.} \\
 \cmidrule(lr){2-3}  \cmidrule(lr){4-5}
 & \textbf{Summer} & \textbf{Winter} & \textbf{Autumn} & \textbf{Summer} & RR\\
\midrule
\multicolumn{6}{l}{\textit{FPFH}\cite{rusu2009fast}}\\
SC2 & 35.87 & 23.88 & 18.26 & 32.80 & 27.63\\
MAC& 46.74 & 27.68 & 33.04 & 43.20 & 36.12 \\
MAC-OP$\diamond$ & 51.09 & 34.26 & 34.78 & 44.80 & 40.53 \\
Ours & 74.74 & 57.79 & 66.09 & 68.80 & \textbf{64.66} \\
\midrule
{\textit{FCGF}\cite{choy2019fully}}\\
SC2 & 64.13 & 41.52 & 60.00 & 64.80 & 54.42 \\
MAC& 75.54 & 42.91 & 71.30 & 73.60 & 61.29 \\
MAC-OP$\diamond$ & 90.76 & 81.66 & 97.39 & 96.00 & 89.06 \\
Ours & 98.37 & 83.05 & 100.00 & 99.20 & \textbf{92.57} \\
\midrule
{\textit{SpinNet}\cite{ao2021spinnet}}\\
SC2 & 96.74 & 82.35 & 97.39 & 95.20 & 92.92 \\
MAC& 98.91 & 87.54 & 100.00 & 100.00 & 94.67 \\
MAC-OP$\diamond$ & 98.91 & 89.97 & 100.00 & 100.00 & 95.65 \\
Ours & 100.00 & 100.00 & 100.00 & 100.00 & \textbf{100.00}  \\
\midrule
\multicolumn{6}{l}{\textit{GeoTransformer}\cite{10076895}}\\
SC2 & 78.26 & 94.46 & 95.65 & 99.20 & 91.30 \\
MAC& 70.11 & 95.85 & 95.65 & 99.20 & 89.76 \\
MAC-OP$\diamond$ & 84.78 & 97.58 & 99.13 & 100.00 & 94.95 \\
Ours & 96.20 & 99.65 & 100.00 & 100.00 & \textbf{98.88} \\

\bottomrule
\end{tabular}
\caption{Performance comparison on ETH dataset.}
\label{tab:ETH_performance}
\end{table}

\subsection{Analytical Experiments}
\subsubsection{The Upper Bound of our approach}
Although the evaluation layer of our PCR method is independent of the quality of initial correspondences, the overall PCR system still relies on generating hypotheses from them. To quantify current limitations, we introduce a new metric $Top_m RR$ as follows: 
\begin{equation}
    Top_m\ RR= \frac{\sum \mathbf{1}(top\ m\ hypotheses)}{\#\ of\ pairs},
\end{equation}
where $\mathbf{1}(top\ m\ hypotheses)$ is defined as 1 if a correct transformation exists among the first m hypotheses, and 0 otherwise.

We analyze $Top_{10}$, $Top_{50}$, and $Top_{100}\ RR$ across descriptors while also assessing the SVC module's role in Tab.\ref{tab:svc_analysis}. In our method, the SVC module primarily enhances the Top-m RR metric by filtering out incorrect transformations. Notably, while PEAL\cite{yu2023peal} demonstrates superior performance without SVC, it underperforms GeoTransformer\cite{qin2022geometric} when SVC is applied. We attribute this to the use of overlap prior, which may reduce the likelihood of generating diverse hypotheses, potentially excluding correct ones.

\begin{table}[!htbp]
\centering
\begin{tabular}{lcccc}
\toprule
& & $Top_{10}$  & $Top_{50}$ & $Top_{100}$ \\
 & RR  &  RR  &  RR  & RR \\
\midrule
\multicolumn{4}{l}{\textit{without SVC\cite{zhang2024svc}}}\\
FCGF & 68.89 & 67.15& 73.66 & 77.37 \\
PREDATOR& 74.40 & 75.96 & 80.12 & 82.93  \\
GeoTr. &84.00 &84.11&86.92& 88.55 \\
PEAL & \textbf{84.56} & \textbf{85.68}& \textbf{87.47}& \textbf{88.65} \\
\midrule
\multicolumn{4}{l}{\textit{with SVC\cite{zhang2024svc}}}\\
FCGF & 71.36 & 74.73 & 78.83 & 80.12 \\
PREDATOR& 78.27 &78.66 &83.77& 86.36 \\
GeoTr. & \textbf{86.97}& \textbf{88.77}& \textbf{91.02} & \textbf{92.31} \\
PEAL & 86.19 & 87.70 & 89.56 & 90.67 \\
\bottomrule
\end{tabular}
\caption{Statistics of Top-m RR on different descriptors}
\label{tab:svc_analysis}
\end{table}

\subsubsection{Performance under different inlier ratio}
\begin{figure}[!htbp]
  \centering
   \includegraphics[width=0.9\linewidth]{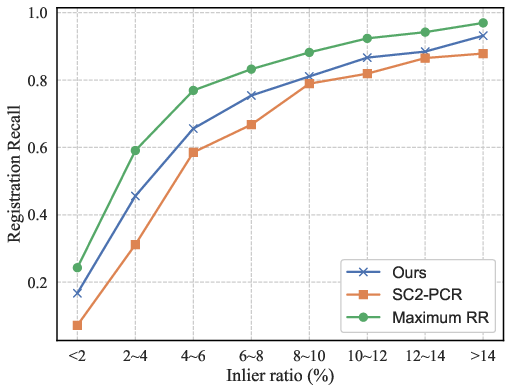}

   \caption{Registration Recall under different inlier ratio}
   \label{fig:RR_vs_IR}
\end{figure}

To investigate the performance of our method under varying inlier rates, we further stratified the inlier rates using FPFH and FCGF on 3DLoMatch and obtained data illustrated in the Fig.\ref{fig:RR_vs_IR}. The Maximum RR means the upper bound with input hypotheses. The figure demonstrates that our method consistently outperforms the original SC2 across all inlier ratios. Notably, the improvement is more pronounced under low inlier ratios ($<$8\%).

\subsubsection{Parameter Experiments}
\begin{figure}[htbp]
    \centering
    \begin{subfigure}[b]{0.48\linewidth} 
        \includegraphics[width=\linewidth]{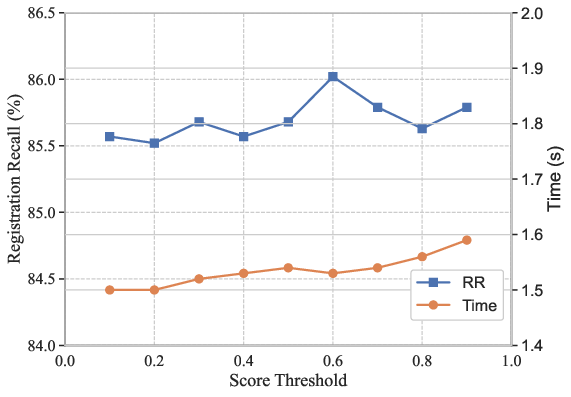} 
        \caption{Score Threshold vs RR} 
        \label{fig:subfig-a}
    \end{subfigure}
    \hfill 
    \begin{subfigure}[b]{0.48\linewidth} 
        \includegraphics[width=\linewidth]{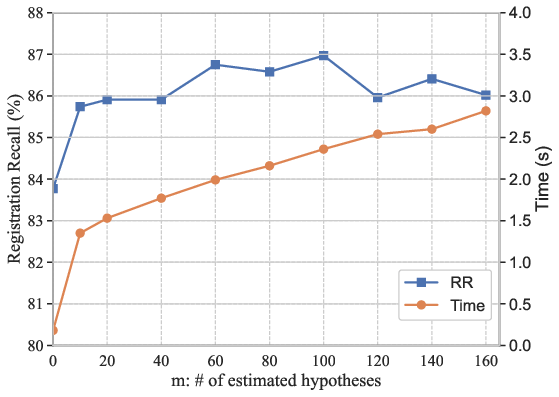} 
        \caption{m vs RR} 
        \label{fig:subfig-b}
    \end{subfigure}
    \caption{Parameter experiments. For the Fig.\ref{fig:subfig-a}, the m is set as 30. For the Fig.\ref{fig:subfig-b}, the Score threshold is set as 0.6.} 
    \label{fig:parameter_RR}
\end{figure}

In this experiment, we primarily investigate the impact of m and Score threshold, as mentioned in Alg.\ref{alg:evaluation}, on Registration Recall (RR) and runtime. As shown in Fig. \ref{fig:parameter_RR}, the score threshold has a slight impact on both RR ($<$0.5) and runtime ($<$0.1s). For the parameter m, we observe an overall increasing trend in RR when m is less than 60, while RR tends to stabilize at larger values. Meanwhile, the runtime is significantly affected by m, showing a steady increase as m grows.

\section{Conclusion}
This work establishes the Decision PCR task as the foundational basis for robust point cloud registration. We propose, for the first time, a data-driven pipeline to address this task. Building upon the Decision PCR model, we introduce a novel PCR method capable of generating uncertainty-aware confidence scores for registration results. This simple yet effective paradigm redefines reliability of the PCR task while offering a versatile tool for future 3D vision systems. Experimental results confirm the superior performance and generalization ability of our approach across both indoor and outdoor benchmarks.

\clearpage









\begin{figure*}[h]
  \centering
  \includegraphics[width=0.95\linewidth]{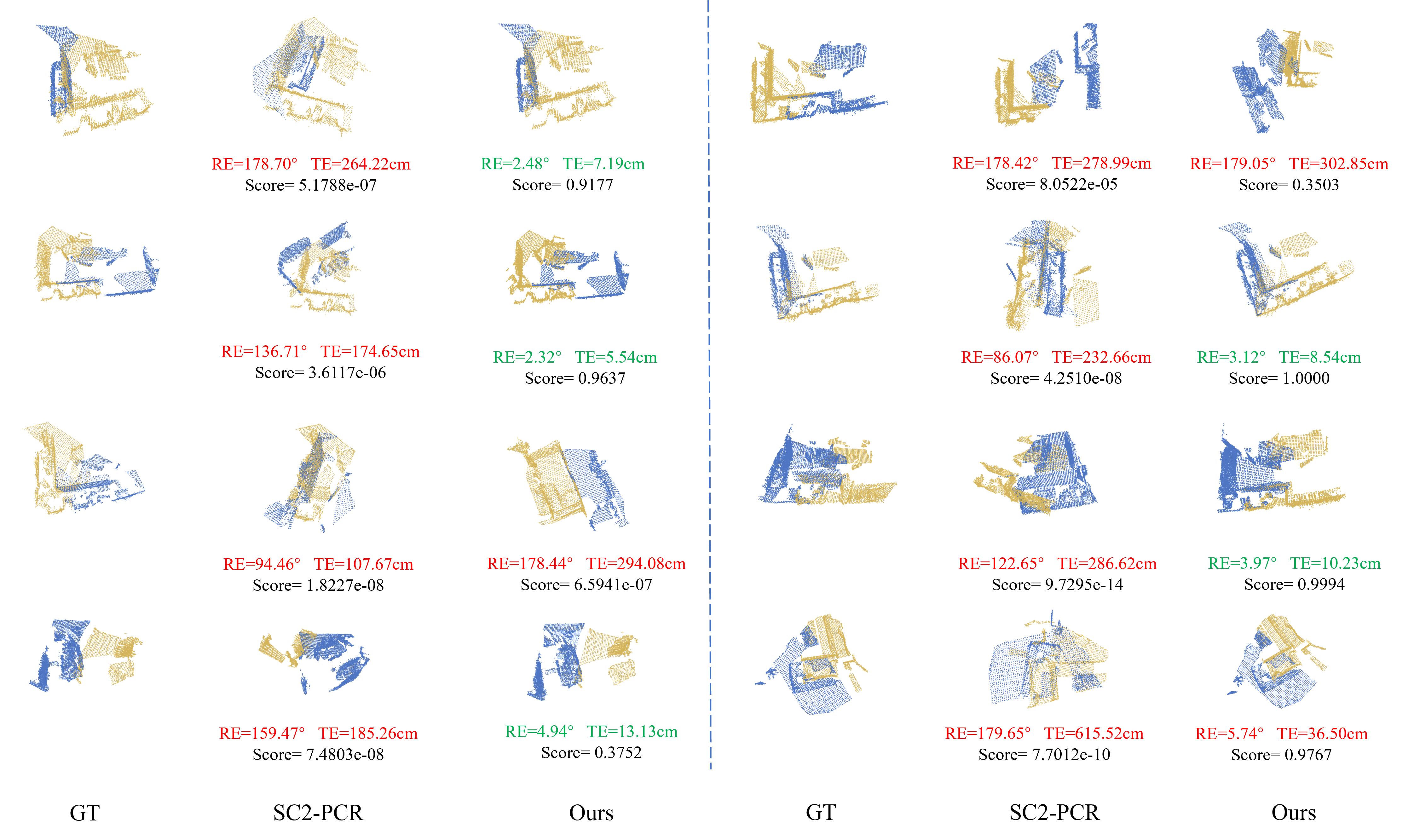}
   \caption{Visualization Results on 3DLoMatch. Red and green colors indicate failed and successful registrations, respectively. The registration score is computed using the Decision PCR model, with higher value indicating a stronger tendency toward positive prediction.}
   \label{fig:visualization_3DLoMatch}
\end{figure*}

\begin{figure*}[h]
  \centering
  \includegraphics[width=0.95\linewidth]{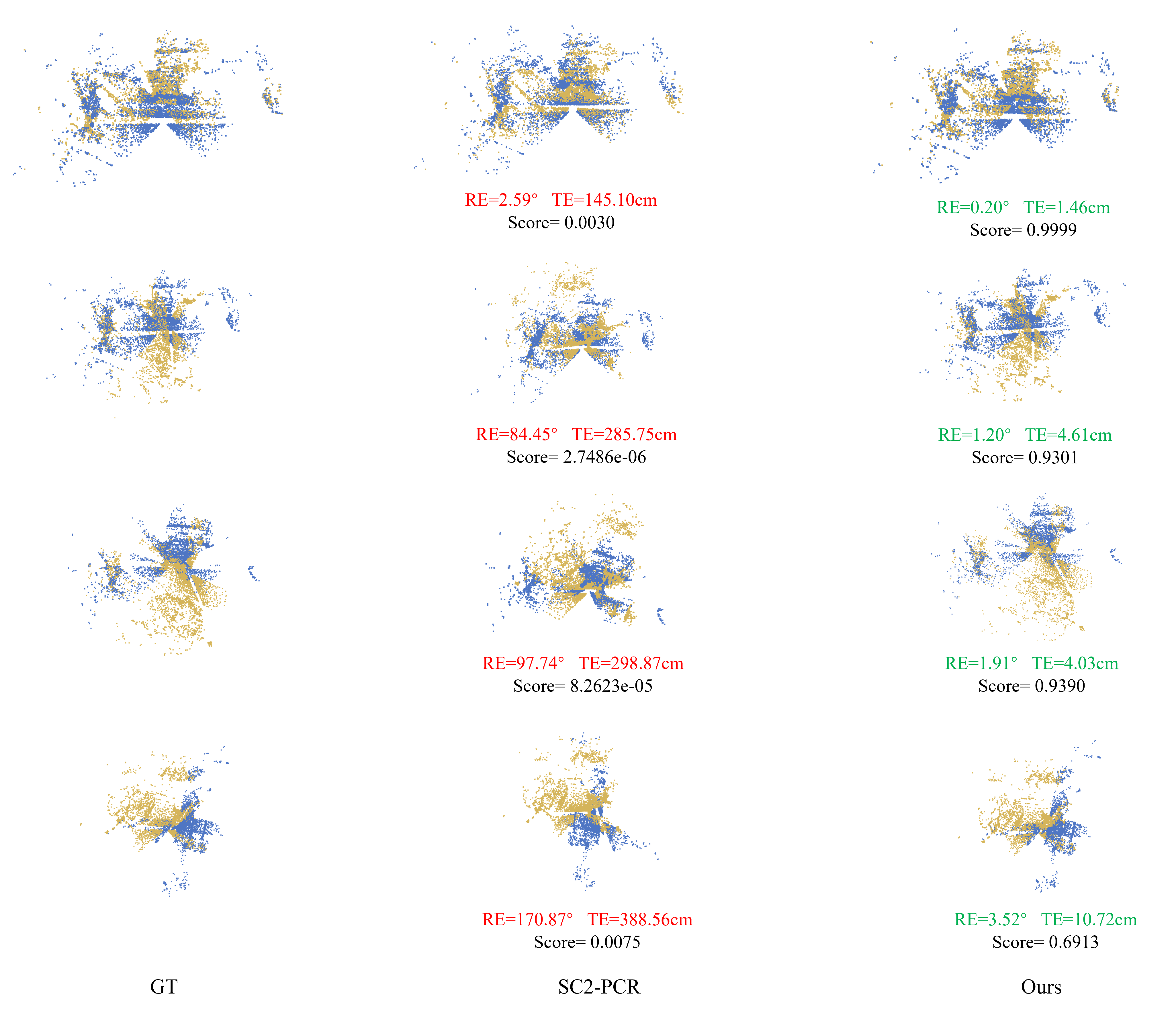}
   \caption{Visualization Results on ETH dataset. Red and green colors indicate failed and successful registrations, respectively. The registration score is computed using the Decision PCR model, with higher value indicating a stronger tendency toward positive prediction.}
   \label{fig:visualization_ETH}
\end{figure*}


\begin{thebibliography}{35}
\providecommand{\natexlab}[1]{#1}
\providecommand{\url}[1]{\texttt{#1}}
\expandafter\ifx\csname urlstyle\endcsname\relax
  \providecommand{\doi}[1]{doi: #1}\else
  \providecommand{\doi}{doi: \begingroup \urlstyle{rm}\Url}\fi

\bibitem[Ao et~al.(2021)Ao, Hu, Yang, Markham, and Guo]{ao2021spinnet}
Sheng Ao, Qingyong Hu, Bo Yang, Andrew Markham, and Yulan Guo.
\newblock Spinnet: Learning a general surface descriptor for 3d point cloud registration.
\newblock In \emph{Proceedings of the IEEE/CVF conference on computer vision and pattern recognition}, pages 11753--11762, 2021.

\bibitem[Aoki et~al.(2019)Aoki, Goforth, Srivatsan, and Lucey]{aoki2019pointnetlk}
Yasuhiro Aoki, Hunter Goforth, Rangaprasad~Arun Srivatsan, and Simon Lucey.
\newblock Pointnetlk: Robust \& efficient point cloud registration using pointnet.
\newblock In \emph{Proceedings of the IEEE/CVF conference on computer vision and pattern recognition}, pages 7163--7172, 2019.

\bibitem[Bai et~al.(2020)Bai, Luo, Zhou, Fu, Quan, and Tai]{bai2020d3feat}
Xuyang Bai, Zixin Luo, Lei Zhou, Hongbo Fu, Long Quan, and Chiew-Lan Tai.
\newblock D3feat: Joint learning of dense detection and description of 3d local features.
\newblock In \emph{Proceedings of the IEEE/CVF conference on computer vision and pattern recognition}, pages 6359--6367, 2020.

\bibitem[Bai et~al.(2021)Bai, Luo, Zhou, Chen, Li, Hu, Fu, and Tai]{bai2021pointdsc}
Xuyang Bai, Zixin Luo, Lei Zhou, Hongkai Chen, Lei Li, Zeyu Hu, Hongbo Fu, and Chiew-Lan Tai.
\newblock Pointdsc: Robust point cloud registration using deep spatial consistency.
\newblock In \emph{Proceedings of the IEEE/CVF Conference on Computer Vision and Pattern Recognition}, pages 15859--15869, 2021.

\bibitem[Barath and Matas(2018)]{barath2018graph}
Daniel Barath and Ji{\v{r}}{\'\i} Matas.
\newblock Graph-cut ransac.
\newblock In \emph{Proceedings of the IEEE conference on computer vision and pattern recognition}, pages 6733--6741, 2018.

\bibitem[Barath et~al.(2019)Barath, Matas, and Noskova]{barath2019magsac}
Daniel Barath, Jiri Matas, and Jana Noskova.
\newblock Magsac: marginalizing sample consensus.
\newblock In \emph{Proceedings of the IEEE/CVF conference on computer vision and pattern recognition}, pages 10197--10205, 2019.

\bibitem[Chen et~al.(2022)Chen, Sun, Yang, and Tao]{chen2022sc2}
Zhi Chen, Kun Sun, Fan Yang, and Wenbing Tao.
\newblock Sc2-pcr: A second order spatial compatibility for efficient and robust point cloud registration.
\newblock In \emph{Proceedings of the IEEE/CVF Conference on Computer Vision and Pattern Recognition}, pages 13221--13231, 2022.

\bibitem[Chen et~al.(2023)Chen, Sun, Yang, Guo, and Tao]{chen2023sc}
Zhi Chen, Kun Sun, Fan Yang, Lin Guo, and Wenbing Tao.
\newblock {SC$^{2}$-PCR++}: Rethinking the generation and selection for efficient and robust point cloud registration.
\newblock \emph{IEEE Transactions on pattern analysis and machine intelligence}, 2023.

\bibitem[Choy et~al.(2019)Choy, Park, and Koltun]{choy2019fully}
Christopher Choy, Jaesik Park, and Vladlen Koltun.
\newblock Fully convolutional geometric features.
\newblock In \emph{Proceedings of the IEEE/CVF international conference on computer vision}, pages 8958--8966, 2019.

\bibitem[Choy et~al.(2020)Choy, Dong, and Koltun]{choy2020deep}
Christopher Choy, Wei Dong, and Vladlen Koltun.
\newblock Deep global registration.
\newblock In \emph{Proceedings of the IEEE/CVF conference on computer vision and pattern recognition}, pages 2514--2523, 2020.

\bibitem[Chum and Matas(2005)]{chum2005matching}
Ondrej Chum and Jiri Matas.
\newblock Matching with prosac-progressive sample consensus.
\newblock In \emph{2005 IEEE computer society conference on computer vision and pattern recognition (CVPR'05)}, pages 220--226. IEEE, 2005.

\bibitem[Fischler and Bolles(1981)]{fischler1981random}
Martin~A Fischler and Robert~C Bolles.
\newblock Random sample consensus: a paradigm for model fitting with applications to image analysis and automated cartography.
\newblock \emph{Communications of the ACM}, 24\penalty0 (6):\penalty0 381--395, 1981.

\bibitem[Geiger et~al.(2013)Geiger, Lenz, Stiller, and Urtasun]{geiger2013vision}
Andreas Geiger, Philip Lenz, Christoph Stiller, and Raquel Urtasun.
\newblock Vision meets robotics: The kitti dataset.
\newblock \emph{The International Journal of Robotics Research}, 32\penalty0 (11):\penalty0 1231--1237, 2013.

\bibitem[Huang et~al.(2021)Huang, Gojcic, Usvyatsov, Wieser, and Schindler]{huang2021predator}
Shengyu Huang, Zan Gojcic, Mikhail Usvyatsov, Andreas Wieser, and Konrad Schindler.
\newblock Predator: Registration of 3d point clouds with low overlap.
\newblock In \emph{Proceedings of the IEEE/CVF Conference on computer vision and pattern recognition}, pages 4267--4276, 2021.

\bibitem[Jiang et~al.(2023)Jiang, Dang, Wei, Xie, Yang, and Salzmann]{jiang2023robust}
Haobo Jiang, Zheng Dang, Zhen Wei, Jin Xie, Jian Yang, and Mathieu Salzmann.
\newblock Robust outlier rejection for 3d registration with variational bayes.
\newblock In \emph{Proceedings of the IEEE/CVF conference on computer vision and pattern recognition}, pages 1148--1157, 2023.

\bibitem[Lee et~al.(2021)Lee, Kim, Cho, and Park]{lee2021deep}
Junha Lee, Seungwook Kim, Minsu Cho, and Jaesik Park.
\newblock Deep hough voting for robust global registration.
\newblock In \emph{Proceedings of the IEEE/CVF international conference on computer vision}, pages 15994--16003, 2021.

\bibitem[Leordeanu and Hebert(2005)]{leordeanu2005spectral}
Marius Leordeanu and Martial Hebert.
\newblock A spectral technique for correspondence problems using pairwise constraints.
\newblock In \emph{Tenth IEEE International Conference on Computer Vision (ICCV'05) Volume 1}, pages 1482--1489. IEEE, 2005.

\bibitem[Pomerleau et~al.(2012)Pomerleau, Liu, Colas, and Siegwart]{pomerleau2012challenging}
Fran{\c{c}}ois Pomerleau, Ming Liu, Francis Colas, and Roland Siegwart.
\newblock Challenging data sets for point cloud registration algorithms.
\newblock \emph{The International Journal of Robotics Research}, 31\penalty0 (14):\penalty0 1705--1711, 2012.

\bibitem[Qin et~al.(2022)Qin, Yu, Wang, Guo, Peng, and Xu]{qin2022geometric}
Zheng Qin, Hao Yu, Changjian Wang, Yulan Guo, Yuxing Peng, and Kai Xu.
\newblock Geometric transformer for fast and robust point cloud registration.
\newblock In \emph{Proceedings of the IEEE/CVF conference on computer vision and pattern recognition}, pages 11143--11152, 2022.

\bibitem[Qin et~al.(2023)Qin, Yu, Wang, Guo, Peng, Ilic, Hu, and Xu]{10076895}
Zheng Qin, Hao Yu, Changjian Wang, Yulan Guo, Yuxing Peng, Slobodan Ilic, Dewen Hu, and Kai Xu.
\newblock Geotransformer: Fast and robust point cloud registration with geometric transformer.
\newblock \emph{IEEE Transactions on Pattern Analysis and Machine Intelligence}, 45\penalty0 (8):\penalty0 9806--9821, 2023.

\bibitem[Quan and Yang(2020)]{quan2020compatibility}
Siwen Quan and Jiaqi Yang.
\newblock Compatibility-guided sampling consensus for 3-d point cloud registration.
\newblock \emph{IEEE Transactions on Geoscience and Remote Sensing}, 58\penalty0 (10):\penalty0 7380--7392, 2020.

\bibitem[Rusu et~al.(2009)Rusu, Blodow, and Beetz]{rusu2009fast}
Radu~Bogdan Rusu, Nico Blodow, and Michael Beetz.
\newblock Fast point feature histograms (fpfh) for 3d registration.
\newblock In \emph{2009 IEEE international conference on robotics and automation}, pages 3212--3217. IEEE, 2009.

\bibitem[Thomas et~al.(2019)Thomas, Qi, Deschaud, Marcotegui, Goulette, and Guibas]{thomas2019kpconv}
Hugues Thomas, Charles~R Qi, Jean-Emmanuel Deschaud, Beatriz Marcotegui, Fran{\c{c}}ois Goulette, and Leonidas~J Guibas.
\newblock Kpconv: Flexible and deformable convolution for point clouds.
\newblock In \emph{Proceedings of the IEEE/CVF international conference on computer vision}, pages 6411--6420, 2019.

\bibitem[Wang et~al.(2023)Wang, Liu, Hu, Wang, Chen, Dong, Guo, Wang, and Yang]{wang2023roreg}
Haiping Wang, Yuan Liu, Qingyong Hu, Bing Wang, Jianguo Chen, Zhen Dong, Yulan Guo, Wenping Wang, and Bisheng Yang.
\newblock Roreg: Pairwise point cloud registration with oriented descriptors and local rotations.
\newblock \emph{IEEE Transactions on Pattern Analysis and Machine Intelligence}, 45\penalty0 (8):\penalty0 10376--10393, 2023.

\bibitem[Wu et~al.(2015)Wu, Song, Khosla, Yu, Zhang, Tang, and Xiao]{wu20153d}
Zhirong Wu, Shuran Song, Aditya Khosla, Fisher Yu, Linguang Zhang, Xiaoou Tang, and Jianxiong Xiao.
\newblock 3d shapenets: A deep representation for volumetric shapes.
\newblock In \emph{Proceedings of the IEEE conference on computer vision and pattern recognition}, pages 1912--1920, 2015.

\bibitem[Xing et~al.(2024)Xing, Lu, Wang, and Xiao]{xing2024efficient}
Xuejun Xing, Zhengda Lu, Yiqun Wang, and Jun Xiao.
\newblock Efficient single correspondence voting for point cloud registration.
\newblock \emph{IEEE Transactions on Image Processing}, 2024.

\bibitem[Yang et~al.(2020)Yang, Shi, and Carlone]{yang2020teaser}
Heng Yang, Jingnan Shi, and Luca Carlone.
\newblock Teaser: Fast and certifiable point cloud registration.
\newblock \emph{IEEE Transactions on Robotics}, 37\penalty0 (2):\penalty0 314--333, 2020.

\bibitem[Yang et~al.(2015)Yang, Li, Campbell, and Jia]{yang2015go}
Jiaolong Yang, Hongdong Li, Dylan Campbell, and Yunde Jia.
\newblock Go-icp: A globally optimal solution to 3d icp point-set registration.
\newblock \emph{IEEE transactions on pattern analysis and machine intelligence}, 38\penalty0 (11):\penalty0 2241--2254, 2015.

\bibitem[Yang et~al.(2024)Yang, Zhang, Wang, Guo, Sun, Wu, Zhang, and Zhang]{yang2024mac}
Jiaqi Yang, Xiyu Zhang, Peng Wang, Yulan Guo, Kun Sun, Qiao Wu, Shikun Zhang, and Yanning Zhang.
\newblock Mac: Maximal cliques for 3d registration.
\newblock \emph{IEEE Transactions on Pattern Analysis and Machine Intelligence}, 2024.

\bibitem[Yu et~al.(2021)Yu, Li, Saleh, Busam, and Ilic]{yu2021cofinet}
Hao Yu, Fu Li, Mahdi Saleh, Benjamin Busam, and Slobodan Ilic.
\newblock Cofinet: Reliable coarse-to-fine correspondences for robust pointcloud registration.
\newblock \emph{Advances in Neural Information Processing Systems}, 34:\penalty0 23872--23884, 2021.

\bibitem[Yu et~al.(2023)Yu, Ren, Zhang, Zhou, Lin, and Dai]{yu2023peal}
Junle Yu, Luwei Ren, Yu Zhang, Wenhui Zhou, Lili Lin, and Guojun Dai.
\newblock Peal: Prior-embedded explicit attention learning for low-overlap point cloud registration.
\newblock In \emph{Proceedings of the IEEE/CVF Conference on Computer Vision and Pattern Recognition}, pages 17702--17711, 2023.

\bibitem[Zeng et~al.(2017)Zeng, Song, Nie{\ss}ner, Fisher, Xiao, and Funkhouser]{zeng20173dmatch}
Andy Zeng, Shuran Song, Matthias Nie{\ss}ner, Matthew Fisher, Jianxiong Xiao, and Thomas Funkhouser.
\newblock 3dmatch: Learning local geometric descriptors from rgb-d reconstructions.
\newblock In \emph{Proceedings of the IEEE conference on computer vision and pattern recognition}, pages 1802--1811, 2017.

\bibitem[Zhang et~al.(2023)Zhang, Yang, Zhang, and Zhang]{zhang20233d}
Xiyu Zhang, Jiaqi Yang, Shikun Zhang, and Yanning Zhang.
\newblock 3d registration with maximal cliques.
\newblock In \emph{Proceedings of the IEEE/CVF Conference on Computer Vision and Pattern Recognition}, pages 17745--17754, 2023.

\bibitem[Zhang et~al.(2024)Zhang, Wang, Huang, Wang, and Feng]{zhang2024svc}
Yaojie Zhang, Weijun Wang, Tianlun Huang, Zhiyong Wang, and Wei Feng.
\newblock Svc: Sight view constraint for robust point cloud registration.
\newblock \emph{Image and Vision Computing}, 152:\penalty0 105315, 2024.

\bibitem[Zhou et~al.(2016)Zhou, Park, and Koltun]{zhou2016fast}
Qian-Yi Zhou, Jaesik Park, and Vladlen Koltun.
\newblock Fast global registration.
\newblock In \emph{Computer Vision--ECCV 2016: 14th European Conference, Amsterdam, The Netherlands, October 11-14, 2016, Proceedings, Part II 14}, pages 766--782. Springer, 2016.

\end{thebibliography}


\end{document}